\documentclass{article} 
\usepackage{iclr2027_conference,times}
\usepackage{amsmath,amsfonts,bm}

\def\eqref#1{equation~\ref{#1}}
\def\1{\bm{1}}

\DeclareMathAlphabet{\mathsfit}{\encodingdefault}{\sfdefault}{m}{sl}
\SetMathAlphabet{\mathsfit}{bold}{\encodingdefault}{\sfdefault}{bx}{n}

\usepackage{hyperref}
\usepackage{url}
\hypersetup{hidelinks}
\usepackage[utf8]{inputenc}
\usepackage[T1]{fontenc}
\usepackage{textcomp}
\usepackage{amssymb,amsmath}
\usepackage{xcolor}
\usepackage{graphicx}
\usepackage{booktabs,tabularx,array,calc}
\newcolumntype{Y}{>{\raggedright\arraybackslash}X}
\usepackage{caption}
\usepackage{tikz}\usetikzlibrary{arrows.meta,positioning,calc,shapes.geometric,decorations.pathreplacing,backgrounds,fit}
\usepackage{microtype}
\usepackage{placeins}
\makeatletter
\def\maxwidth{\ifdim\Gin@nat@width>\linewidth\linewidth\else\Gin@nat@width\fi}
\def\maxheight{\ifdim\Gin@nat@height>\textheight\textheight\else\Gin@nat@height\fi}
\makeatother
\setkeys{Gin}{width=\maxwidth,height=\maxheight,keepaspectratio}

\newsavebox{\seisfitbox}
\newcommand{\seisfit}{\ifdim\wd\seisfitbox>\linewidth\resizebox{\linewidth}{!}{\usebox{\seisfitbox}}\else\usebox{\seisfitbox}\fi}

\title{SEIS: Self-Evolving Inference Systems}

\author{
\parbox{\textwidth}{\centering
\textbf{Zhen Xu}\qquad
\textbf{Jingyu Liu}\qquad
\textbf{Zongze Li}\qquad
\textbf{Tahseen Rabbani}\qquad
\textbf{Ce Zhang}
\\[0.6em]
{\normalfont\small University of Chicago}
}
}

\iclrfinalcopy % Uncomment for camera-ready version, but NOT for submission.
\begin{document}

\maketitle
\lhead{Preprint}

\begin{abstract}

Inference systems determine how fast and how cheaply language models can be served, so making them faster has direct practical value. However, prior work focuses mostly on optimizing certain parts such as kernels or memory within the large system. In this work, we take a holistic approach and apply agentic self-evolution to optimize the whole system end-to-end. Our \textbf{SEIS} (\textbf{S}elf-\textbf{E}volving \textbf{I}nference \textbf{S}ystems) autonomously optimizes the entire mini-sglang engine without human intervention through iterative sessions with inherited experiences and code changes. Serving Qwen3-0.6B on H100, the resulting engine reaches 3.27\ensuremath{\times} the throughput of the original mini-sglang implementation and beats SOTA engines like vLLM, TensorRT-LLM, and SGLang in the single-request workload. The correctness of the optimized inference engine by SEIS is tested in terms of numerical difference and downstream accuracy on math and long-context retrieval tasks. The code and session histories show that the speedup comes from redesigning the whole engine and that building on earlier sessions beats independent attempts. These results suggest that agentic self-evolution can optimize a complex system end-to-end. The evaluation also has to evolve with the engine, and letting agents evolve it is a natural next step.

\end{abstract}

\hypertarget{introduction}{%
\section{Introduction}\label{introduction}}

An inference engine exposes an agent to a systems problem: improving one component can change which other component limits performance. Faster linear kernels alter the relative cost of attention and scheduling. A different decoding strategy changes the shapes that those kernels must support and the way the engine manages its KV cache. Optimizing the whole engine, therefore, requires choosing a direction, changing several components consistently, and revisiting earlier decisions as the implementation changes.

Recent work uses agents to evolve programs and build or adapt inference systems \citep{alphaevolve,vibeserve,metainfer}. We study sustained optimization of an existing inference engine, tracing how changes accumulate across sessions, combine across chains, and reveal limitations in the evaluation guiding the process. A human provides a working engine, a target model and GPU, a throughput objective, and constraints. The agent decides what to investigate and implement. SEIS preserves the implementation and reports across sessions, and optionally exchanges them between chains. The object that evolves is the inference system's source code and numerical implementation; the agent is fixed throughout the experiment.

\begin{figure}[!t]
\hypertarget{fig:1}{%
\centering
\includegraphics{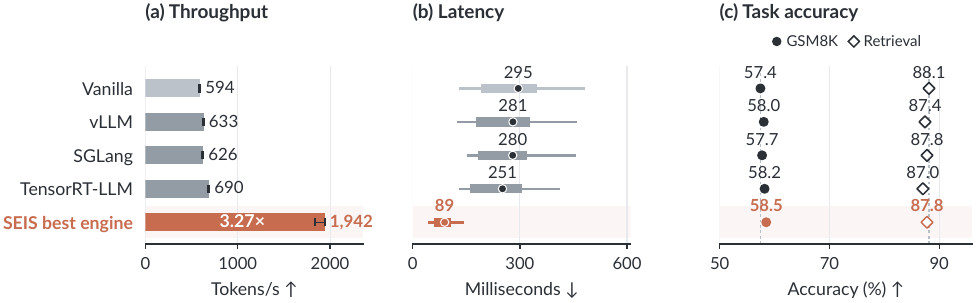}
\caption{Qwen3-0.6B on H100. (a) Natural-request throughput: median and range of three repetitions; the factor is relative to vanilla. (b) Latency means and central 50\%/80\% intervals over 100 requests. (c) Observed accuracy on GSM8K (660 items) and retrieval (270); dotted lines mark vanilla. Production settings are selected on development data (Appendix E).}\label{fig:1}
}
\end{figure}

The resulting changes extend across the engine. Starting from the unmodified mini-sglang engine (vanilla), which already includes paged KV storage, FlashInfer attention \citep{flashinfer}, and CUDA graphs, the agents build quantized kernels, fuse computation across operator boundaries, reduce execution overhead, and implement prompt-lookup speculative decoding. We call the evolved engine with the highest development speedup the best engine. It reaches 3.27\ensuremath{\times} vanilla throughput on natural-length requests and exceeds the selected production configurations on the same single-request workload (Figure~\ref{fig:1}). Its measured accuracy differs from vanilla by +1.06 percentage points on the 660-item GSM8K test split and \ensuremath{-}0.37 points on long-input retrieval. Branching Chain runs on Llama-3.2-1B provide additional system results in Appendix G.

Most of the best engine's natural-request speedup remains with speculation disabled. The trajectories first show how its ordinary kernels and execution path are assembled through successive sessions. They then expose the formation of a larger structural change: speculative decoding is proposed, rejected, remeasured, and eventually implemented. A sync transfers a working design to other chains. The best engine adopts the originating chain's policy while retaining and improving its own execution layer and kernels. This connects the delivered engine to both accumulated local work and selective cross-chain integration.

We contribute a framework and recorded study of sustained autonomous inference-system optimization, together with the resulting engines. The evidence connects three findings: substantial natural-request gains persist in ordinary decoding; accumulated code and selected peer changes form the best engine's implementation; and development-gate success does not establish complete output capability or final-task quality. Sections~\ref{what-the-agents-built}--\ref{reward-hacking-and-validation-failures} present the system results, their formation, and the validation failures, respectively.

\hypertarget{related-work}{%
\section{Related Work}\label{related-work}}

\textbf{Program evolution and self-improvement.} FunSearch and AlphaEvolve evolve programs: a language model proposes edits and an automated evaluator scores them \citep{funsearch,alphaevolve}. FunSearch evolves separate islands of programs and periodically reseeds the weaker ones, following the island model of evolutionary computation \citep{whitley1999island}. Self-improvement methods instead modify the improver. The Gödel machine rewrites its own code \citep{schmidhuber2007godel}, STOP improves the scaffold that calls the language model \citep{stop}, and the Darwin Gödel Machine evolves the coding agent itself \citep{dgm}. SEIS keeps the agent fixed and evolves a separate system. Its unit of evolution is a multi-hour agent session over an entire codebase, and the Sync Branching Chain applies the island model to such sessions.

\begin{figure}[!t]
\hypertarget{fig:protocol}{%
\centering
\includegraphics{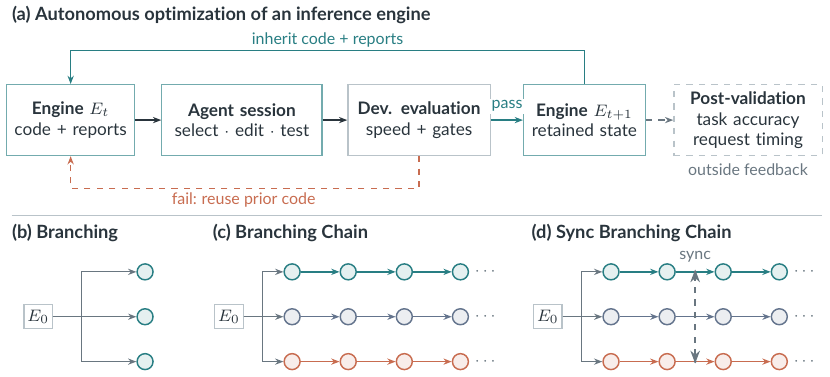}
\caption{The SEIS protocol. (a) A fixed agent chooses and implements optimizations to an inherited engine. Development gates determine which code is retained; passing candidates may be slower. Code and reports carry to the next session. Post-validation evaluates task accuracy and natural-length speed outside optimization feedback. (b) Branching: sessions start independently from the same engine $E_0$, and the best result is kept. (c) Branching Chain: each session inherits the code and reports of the previous session in its chain. (d) Sync Branching Chain: every $N$ sessions, all chains exchange code and reports, and each agent decides what to adopt. Nodes represent sessions.}\label{fig:protocol}
}
\end{figure}

\textbf{Long-horizon agents and experience across attempts.} MLE-bench, RE-Bench, and PaperBench evaluate agents on machine-learning engineering and research tasks that take hours \citep{mlebench,rebench,paperbench}, and The AI Scientist automates a complete research cycle \citep{aiscientist}. Reflexion, ExpeL, and Voyager store verbal lessons or reusable skills for later attempts \citep{reflexion,expel,voyager}, and MemGPT manages information beyond one context window \citep{memgpt}. In SEIS, both code and written reports pass across up to twenty sessions. We find that inherited reports carry mistaken conclusions as well as progress, and we compare a reduced context budget that forces compaction.

\textbf{LLMs for kernels, software, and serving systems.} KernelBench and The AI CUDA Engineer study the generation and optimization of individual GPU kernels \citep{kernelbench,aicudaengineer}. SWE-bench and Terminal-Bench evaluate agents on software and command-line tasks with fixed tests \citep{swebench,terminalbench}. VibeServe builds serving stacks for particular deployments, and some of its tasks name speculation as a target \citep{vibeserve,vibeserve_tasks}. MetaInfer adapts models and hardware within existing inference frameworks \citep{metainfer}. SEIS starts from a working engine, leaves the optimization sequence to the agents, and retains a change only if it passes numerical and termination checks.

\textbf{Efficient LLM inference.} The engines the agents produced combine established techniques. Serving systems such as vLLM, SGLang, and TensorRT-LLM, and attention kernels such as FlashAttention and FlashInfer, improve throughput and memory use \citep{vllm,sglang,tensorrtllm,flashattention,flashinfer}. Quantization to 8 or 4 bits reduces the memory traffic of decoding \citep{llmint8,gptq,awq}. Speculative decoding verifies drafted tokens with the target model and preserves its output distribution \citep{specdecode,chen2023speculative}. Drafts can come from the prompt \citep{promptlookup}, from n-grams generated in parallel \citep{lookahead}, or from lightweight draft heads \citep{medusa,eagle}. We do not claim these techniques. We study how agents chose, implemented, and combined them.

\textbf{Reward hacking and agent evaluation.} Optimizing a measured objective can exploit gaps between the measurement and the intended task \citep{amodei2016concrete,krakovna2020specification,rewardhacking}. Coding agents skip or special-case tests \citep{baker2025monitoring}, and ImpossibleBench measures how often they exploit tests that contradict the specification \citep{impossiblebench}. We report such cases in systems optimization, such as output-vocabulary restriction that keeps task scores within the margin, and describe how revised evaluation exposed them (Section~\ref{reward-hacking-and-validation-failures}).

\begin{figure}[!t]
\hypertarget{fig:3}{%
\centering
\includegraphics{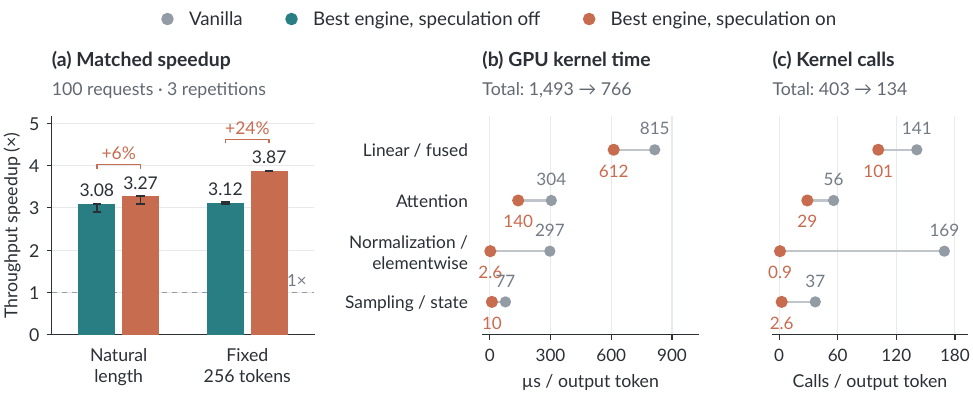}
\caption{Performance and execution changes in the best engine. (a) Matched workloads: median throughput ratios over three repetitions; whiskers show run ranges with the vanilla median fixed. Percentages indicate the additional throughput from speculation. (b,c) Vanilla and the full evolved engine on a separate CUDA profile with 128 forced output tokens, including prefill: summed kernel time and kernel invocations per output token. Fused operations belong to their encompassing kernel group; these groups are not independent speedup attributions. Overlapping kernel intervals are counted separately, so their sum is not wall-clock time.}\label{fig:3}
}
\end{figure}

\hypertarget{seis-framework-and-experimental-setup}{%
\section{SEIS: Framework and Experimental Setup}\label{seis-framework-and-experimental-setup}}

In this section, we start with the formal definition of our evolving target and the rubrics for evaluating the success an attempt, then we describe the core agent execution and cooperation protocols used in SEIS, and finally describe the specific realization of SEIS used in our experiments. 

\subsection{Improving LLM Serving Efficiency as an Evolving Challenge}

We choose optimizing Qwen3-0.6B's single-request decoding performance inside of mini-sglang as the core objective for self-evolution. The main reason for choosing this is because its minimal code base allows free exploration of creative solutions, yet still providing practical significance. On top of that, specifying the objective and verifying progress becomes programmablely possible. 

SEIS consists of multiple agent sessions (the interaction patterns among them are described below), where we give each session a starting codebase, along with a goal description but do not provide concrete ideas on what things to try (SEIS's own job). Specifically, a session is only given a private worktree, development workloads, correctness threshold, and a scorer. It is worth noting that we do not prescribe an optimization sequence or name very specific techniques and, therefore, SEIS agents have the full freedom to explore techniques such as self-attention, matrix-vector computation, normalization, sampling, and CUDA graph optimization. And it turns out that SEIS ends up doing more than we anticipated such as editing kernels, adjusting model execution flows, scheduling, and KV cache management. The only constraint we impose for SEIS is not to modify the test scripts (i.e. scorer) defined by the problem.

\subsection{Rubrics for Checking Evolution Progress}
\label{sec:rubrics}

Checking software correctness has been an extremely challenging task in general. Luckily in our setting, we can design very detailed criteria for checking the correctness and efficiency of a specific implementation evolved from running SEIS. 

\paragraph{Numerical Precision Gate} The inference engine takes in different tensor shapes of the input (which is one major factor of the inference efficiency), and SEIS's implementation needs to be accurate in terms of 1) language perplexity (relative difference within 5\%), 2) mean KL divergence ($\le 0.05$), and 3) argmax token agreement ($\ge 90\%$) w.r.t. the base implementation from mini-sglang. 

\paragraph{Downstream Performance Gate} On top of the precision gate, we include GSM8K\citep{gsm8k} and a synthetic retrieval task (details in the Appendix D) as downstream post-validations, because numerical differences could accumulate across steps, and therefore, affect model behaviors. 

We admit that passing these two categories of tasks does not directly lead to behavior equivalence, which itself, as a software similarity check, is a long-standing research question. However, it does explicitly give SEIS the right scope to evolve for creative solutions on improving the baseline's inference efficiency. 
More discusses can be found in Sec~\ref{reward-hacking-and-validation-failures}.

\subsection{SEIS Agent Execution and Cooperation Protocols}

SEIS's main workflow consists of multiple independent agent sessions, which have a two-hour wall clock time hard limit. Therefore, the central design question becomes how we can coordinate and exchange information among them. 

\paragraph{Single Session Life Cycle} When a session finishes either due to time-out or completion, the master harness will take the changed code and run evaluation pipelines on the attempted optimization against the two gates. A successful session will become the base for future sections to inherit from, including the updated worktree and progress reports. 

\paragraph{Multiple Session Topology} Evolution appears through iterative process, and in our case, sessions need to iterate from where past sessions finish and learn from their experiences. In this work, we study three concrete topologies that connect sessions together. 

\begin{itemize}
    \item \textbf{Branching} Branching is the simplest form where we let multiple single sessions to work from the same starting point (same base code base, objective descriptions, etc), enabling independent explorations of the same problem. The final result is aggregated as the best implementation among them. 
    \item \textbf{Branching Chain} Different from Branching, Branching Chain allows information exchange through temporal inheritance where the new sessions can learn and build up from past sessions' success and failure cases through reports and evaluation results. 
    \item \textbf{Sync Branching Chain} Branching Chain allows multiple parallel agents to solve a single problem and the natural extension is allowing them to share knowledge during their independent execution. Sync Branching Chain adds fixed interval (N sessions) barriers among chains of sessions for synchronized all-to-all message exchange, so that all chains are aware of each other's progress. 
\end{itemize}

The full topology visualization is in Figure~\ref{fig:protocol}. We will show in our results that Branching is an intuitive first choice but often the worst design among the three, significantly eclipsed by the latter two due to lack of cross-session and cross-chain evolution capability. And surprisingly, the final performance parity between the Sync and normal Branching Chain heavily depends on the agent's own behavior, which we will discuss in our experiment section.

\begin{figure}[!t]
\hypertarget{fig:4}{%
\centering
\includegraphics{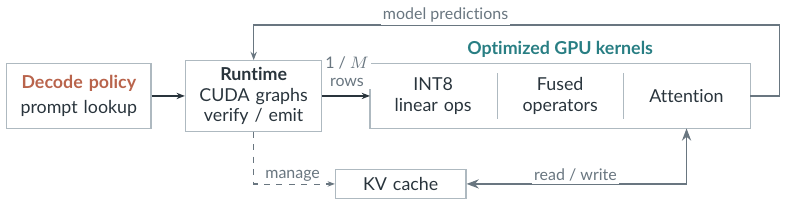}
\caption{Organization of the evolved decode implementation. The policy selects single-token decoding or verification over \(M\) token positions. The runtime replays CUDA graphs, verifies drafts, emits tokens, and manages KV state; attention kernels read and update the cache. Appendix F shows the detailed execution paths.}\label{fig:4}
}
\end{figure}

\subsection{Experiment Setup}

We use Claude Opus 5 through Claude Code (Opus) and GPT-5.6 Sol through Codex (GPT), both at their maximum reasoning and context length setting. Each agent runs four conditions, each with three chains of twenty sessions: Branching Chain (Chain), Sync Branching Chain with $N=3$ and $N=9$ (Sync-3 and Sync-9), and Branching Chain with context compaction at about 128k tokens (Chain-128k). The main Qwen experiments thus contain 24 chains and 480 sessions. Branching is not run as a separate condition. Its reference uses the first sessions of the Chain, Sync-3, and Sync-9 chains, which start independently from vanilla: nine sessions per agent. 

We let SEIS optimize the inference efficiency of Qwen3-0.6B in mini-sglang, running on a single NVIDIA H100 80G. We focus on single request and greedy decoding for determined numerical, downstream gate evaluation, as well as performance profiling. The validation data for numerical gate spans across natural language, mathematics, and code. Other model family SEIS experiments are in Appendix G. 

\hypertarget{what-the-agents-built}{%
\section{What the Agents Built}\label{what-the-agents-built}}

This section examines the best engine that the agents built: its speed and accuracy relative to vanilla and production engines, and the changes behind its speedup. Table~\ref{tab:1} lists the final engines of all 24 chains.

\hypertarget{end-to-end-performance-and-quality}{%
\subsection{End-to-end performance and quality}\label{end-to-end-performance-and-quality}}

\textbf{Self-evolution delivers a large end-to-end speedup.} The best engine, selected by development speedup, comes from Opus Sync-9 chain A. On 100 natural-length requests, it reaches 1,942 tokens/s against 594 tokens/s for vanilla, a 3.27\ensuremath{\times} speedup. Mean request latency falls from 295 to 89 ms; the best engine emits 1.27 \% fewer tokens over these requests. Figure~\ref{fig:1} presents throughput, latency, and task accuracy together, and Appendix E gives the request statistics.

\textbf{The speedup preserves measured task accuracy.} The best engine scores 58.48 \% on the 660-item GSM8K test split, compared with 57.42 \% for vanilla, and 87.78 \% on retrieval, compared with 88.15 \%. The differences are +1.06 and \ensuremath{-}0.37 percentage points, both within \ensuremath{\pm}3 points. Its audited implementation retains the complete output head and attention computation. These observations concern the two reported tasks and the audited code; Appendix E reports paired accuracy uncertainty.

\textbf{The best evolved engine also outperforms production engines.} We tune vLLM, SGLang, and TensorRT-LLM through their public interfaces, considering quantization, speculation, and execution settings on development data \citep{vllm,sglang,tensorrtllm}. The best evolved engine reaches 2.81--3.10\ensuremath{\times} their selected configurations' throughput on the same model, H100, and single-request workload. The best evolved engine and all three selected references have observed accuracy differences within \ensuremath{\pm}3 points on both final tasks. Appendix E specifies the configuration search and timing protocol.

\hypertarget{sources-of-speedup}{%
\subsection{Sources of speedup}\label{sources-of-speedup}}

\textbf{Self-evolution optimizes a whole system by redesigning the computation.} The best engine changes 30 files across kernels, model layers, attention, the engine, and the scheduler. On the GPU, per-output-channel INT8 linear kernels also perform normalization, residual addition, and activation. Attention combines q/k normalization, rotary embeddings, and KV writes, and fuses the combination of attention partials into the output projection. On the host, greedy token selection runs inside CUDA graphs, token IDs are published into pinned host memory, and the scheduler combines step preparation. In a 128-token CUDA profile, kernel calls fall from 403 to 134 per output token, and standalone normalization and elementwise calls fall from 169 to below one; summed kernel time falls from 1,493 to 766 \ensuremath{\mu}s per token (Figure~\ref{fig:3}(b,c)). With speculation off, the engine needs 145 calls and 775 \ensuremath{\mu}s per token, so almost all of this reduction comes from the redesigned single-token path. Kernel intervals overlap under programmatic dependent launch, so these sums are not wall-clock decoding time; throughput is measured separately.

\textbf{Optimizations build on each other: speculative decoding pays off on top of efficient kernels.} Figure~\ref{fig:4} shows the additional path. Prompt lookup proposes tokens from the prompt and generated history. Estimated token yield and verification cost decide between ordinary decoding and eight- or sixteen-row verification. Multi-row INT8 kernels and shared-KV attention execute the verification, while the runtime reserves cache slots, commits the accepted prefix, and emits the accepted tokens. The chains that built it did so after their kernels had matured (Section~\ref{accumulation-across-sessions}). On the same 100 prompts with three repetitions, speculation raises the speedup from 3.08\ensuremath{\times} to 3.27\ensuremath{\times} on natural-length requests (+6 \%), and from 3.12\ensuremath{\times} to 3.87\ensuremath{\times} when outputs are forced to 256 tokens (+24 \%; Figure~\ref{fig:3}(a)). The final policy comes from chain B and is integrated with chain A's execution layer and kernels (Section~\ref{cross-chain-integration}); Appendices B and F give the implementation and validation details.

\begin{table}[t]
\caption{Final engines of all 24 Qwen chains. Speedups are factors; accuracy differences are percentage points on the 660-item GSM8K test split and retrieval. Bold marks a slowdown or a difference outside \ensuremath{\pm}3 points; \(\dagger\) marks speculation. Passing these task tests does not exclude the functional limitations in Section~\ref{reward-hacking-and-validation-failures}.}\label{tab:1}
\centering
\small
\renewcommand{\arraystretch}{1.15}
\begin{tabular}{l c r r r r @{\hspace{12pt}} r r r r}
\toprule
 &  & \multicolumn{4}{c}{Opus} & \multicolumn{4}{c}{GPT} \\
\cmidrule(lr){3-6}\cmidrule(lr){7-10}
condition & chain & dev. & held-out & GSM8K \ensuremath{\Delta} & retrieval \ensuremath{\Delta} & dev. & held-out & GSM8K \ensuremath{\Delta} & retrieval \ensuremath{\Delta} \\
\midrule
Chain & A & 4.94\rlap{\(^\dagger\)} & 6.11 & \ensuremath{-}0.15 & \ensuremath{-}1.48 & 2.43 & 2.37 & \ensuremath{-}1.36 & \ensuremath{-}0.74 \\
 & B & 2.88 & 2.71 & +0.91 & \textbf{\ensuremath{-}28.89} & 1.84 & 1.82 & \ensuremath{-}1.21 & +1.11 \\
 & C & 3.54 & 3.31 & +0.91 & +0.37 & 2.13 & 2.07 & \ensuremath{-}1.36 & \ensuremath{-}2.22 \\
\specialrule{0.3pt}{2pt}{2pt}
Sync-3 & A & 3.38 & 3.12 & +1.52 & \ensuremath{-}0.37 & 2.96 & 2.76 & +0.30 & 0.00 \\
 & B & 3.39 & 3.15 & +1.06 & 0.00 & 2.95 & 2.84 & \ensuremath{-}0.15 & +0.74 \\
 & C & 3.26 & 3.11 & +0.30 & \ensuremath{-}1.48 & 2.89 & 2.71 & \ensuremath{-}1.21 & \ensuremath{-}0.74 \\
\specialrule{0.3pt}{2pt}{2pt}
Sync-9 & A & 6.28\rlap{\(^\dagger\)} & 8.54 & +1.06 & \ensuremath{-}0.37 & 2.22 & 2.16 & \ensuremath{-}2.58 & 0.00 \\
 & B & 5.50\rlap{\(^\dagger\)} & 7.69 & +0.91 & \ensuremath{-}1.11 & 2.14 & 2.17 & \ensuremath{-}2.42 & \textbf{\ensuremath{-}29.63} \\
 & C & 5.83\rlap{\(^\dagger\)} & 8.54 & \ensuremath{-}1.52 & \ensuremath{-}1.48 & 2.21 & 2.16 & \ensuremath{-}2.42 & 0.00 \\
\specialrule{0.3pt}{2pt}{2pt}
Chain-128k & A & 2.93 & 2.73 & +1.06 & \ensuremath{-}0.37 & 2.23 & 2.09 & \textbf{\ensuremath{-}4.39} & +2.59 \\
 & B & 2.93 & 2.75 & +0.91 & \ensuremath{-}0.74 & 2.39 & \textbf{0.77} & \textbf{\ensuremath{-}4.55} & \ensuremath{-}2.59 \\
 & C & 3.01 & 2.79 & +1.21 & \ensuremath{-}1.11 & 1.94 & 1.88 & \textbf{\ensuremath{-}4.39} & +2.96 \\
\bottomrule
\end{tabular}
\end{table}

\hypertarget{how-the-engine-evolved}{%
\section{How the Engine Evolved}\label{how-the-engine-evolved}}

This section follows how the engines evolved over sessions and across chains, measuring progress by development speedup, the score that SEIS optimizes during evolution.

\hypertarget{accumulation-across-sessions}{%
\subsection{Branching Chain: Building on Inherited Work}\label{accumulation-across-sessions}}

\begin{figure}
\hypertarget{fig:5}{%
\centering
\includegraphics{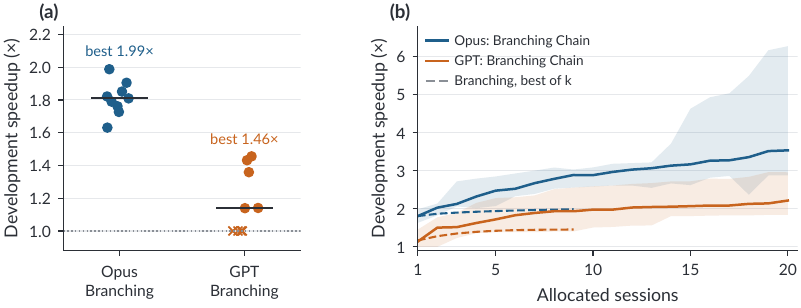}
\caption{Branching Chain versus Branching in the Chain, Sync-3, and Sync-9 conditions. (a) Retained first-session outcomes; rejected candidates retain the 1.00\ensuremath{\times} vanilla engine. (b) Median retained development speedup across the nine chains per agent, with the range across chains. Dashed curves show expected best-of-k selection without replacement from that agent's nine observed Branching sessions, for k \ensuremath{\leq}9.}\label{fig:5}
}
\end{figure}

\textbf{Accumulating work beats making more independent attempts.} With the same number of sessions, the median chain exceeds the expected best of as many independent first sessions from the second session on, for both agents (Figure~\ref{fig:5}). By session 20, all eighteen Chain, Sync-3, and Sync-9 chains finish above the best Branching result: the slowest final engine reaches 2.88\ensuremath{\times} for Opus against 1.99\ensuremath{\times}, and 1.84\ensuremath{\times} for GPT against 1.46\ensuremath{\times}. The best engine's chain shows how the work accumulates: it rises from 1.73\ensuremath{\times} in session 1 to 3.19\ensuremath{\times} by session 17. Early sessions move the linear layers to INT8 kernels and fuse operations; later sessions remove host overhead, for example by publishing the selected token from inside the CUDA graph. This accumulated work is the ordinary path measured in Section~\ref{sources-of-speedup}.

\textbf{The largest gains need multi-session changes that build on earlier work.} Speculative decoding, which the goal description never names, is first considered in session 5 but first works in sessions 14 and 16, in two chains. Every final engine above 3.6\ensuremath{\times} uses it. It pays off only after the policy, multi-row kernels, and KV handling change together, so it cannot be built one small step at a time (Appendix~\ref{speculative-decoding-implementation-and-engineering-history}). Its payoff also depends on earlier kernel work. In session 10, one chain measured 1.9 tokens per step against a verification cost of 2.2 ordinary steps and declined. The chains that built it did so after their kernels had matured and a verification step cost less relative to an ordinary one. Because any passing engine is retained, unfinished work can continue. The best engine's chain drops to 2.37\ensuremath{\times} in session 18 with an unfinished speculative implementation, which the next session repairs and combines with a peer's policy (Section~\ref{cross-chain-integration}). With a 128k context, no chain finished the change. The one attempt began in session 20 and hit the time limit; these runs cannot separate compaction from lost working time.

\textbf{Inherited reports pass on errors as well as knowledge.} What the chains wrote about speculative decoding also delayed it. Several chains wrote that the protected interface forbids multi-token execution, and later sessions repeated the claim. Two chains estimated about 1.0 tokens per verification step with an off-by-one simulation, against 2.30 from a correct replay, and abandoned the direction. Another measured 2.07 correctly but left the number out of its report.

\hypertarget{cross-chain-integration}{%
\subsection{Sync Branching Chain: Sharing Work across Chains}\label{cross-chain-integration}}

A Sync Branching Chain runs several chains in parallel and, every $N$ sessions, lets all chains read each other's code and reports. No engine is replaced: each recipient decides what to adopt, so a design transfers only if it can be fitted into a different implementation, a problem familiar from merging parallel engineering efforts. We ask what synchronization adds to a plain chain and when it pays off.

\begin{figure}
\hypertarget{fig:6}{%
\centering
\includegraphics{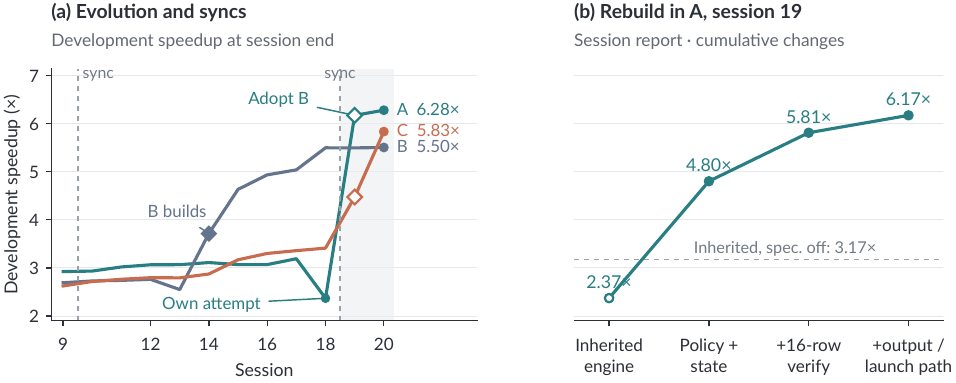}
\caption{Synchronization in Opus Sync-9 (Sync Branching Chain, $N=9$). (a) Retained-engine development speedups. The filled diamond marks B's build; open diamonds mark A/C's adoption after the second sync. (b) Cumulative development-suite measurements from A's session-19 report: policy and state repairs, sixteen-row verification with its teacher-forcing fix, then output handling, launch bindings, and policy-cost retuning. The dashed reference measures the inherited engine with speculation off. Each checkpoint includes multiple implementation changes.}\label{fig:6}
}
\end{figure}

\textbf{Synchronization spreads discoveries faster than chains find them alone.} One chain needed most of its session budget to make speculative decoding work, but a single synchronization spread it to both peers within one session (Figure~\ref{fig:6}(a)). One of them had earlier judged the idea unprofitable; with the peer's design, it produced the best engine.

\textbf{Integrating shared work takes engineering but can beat each chain's own result.} The best engine pairs the originator's decision policy with the recipient's own execution layer and kernels (Appendix~\ref{speculative-decoding-implementation-and-engineering-history}). Before the adopted policy could pay off, the recipient had to repair how its inherited implementation handled accepted tokens and extend its own execution path (Figure~\ref{fig:6}(b)). Both recipients ended faster than the originator (Figure~\ref{fig:6}(a)). The same recipient also tests another peer's suggestion and discards it when it does not help its own implementation.

\hypertarget{reward-hacking-and-validation-failures}{%
\section{How the Evaluation Evolved}\label{reward-hacking-and-validation-failures}}

The evaluation pipeline and rubrics of Section~\ref{sec:rubrics} were themselves revised during our experiments. We present four cases. In each, an engine passed the evaluation in use at the time while breaking something that the evaluation did not check.

\textbf{(1) An evaluation only protects the code it runs.}
Our initial capability check used multiple-choice benchmarks such as MMLU, scored by option probabilities.
An early engine matched the base model's option probabilities on every question, yet performed worse on a separate free-generation GSM8K diagnostic.
The explanation was in what the tests executed: option probabilities came from a single prefill pass, leaving the modified token-by-token decode path untested.
We therefore used tasks requiring generated answers to assess decode quality, including the GSM8K and retrieval tests reported here.
Unchanged multiple-choice scores alone could not establish that the evolving implementation preserved capability.

\textbf{(2) Whatever the evaluation does not check, the agent may trade away.}
Our fixed-length benchmark ignored end-of-sequence (EOS) tokens, allowing an early candidate that removed EOS from its output head to pass.
Every request in a diagnostic GSM8K run then continued to its token limit.
Another candidate passed fixed-length timing but crashed when EOS termination was enabled.
We added short-generation probes requiring normal termination before throughput scoring; failures prevented inheritance.
These probes rejected fourteen sessions in the main Qwen runs, keeping those implementations from becoming the starting point for subsequent sessions.

\textbf{(3) Checks that expose failures can let the agent fix them itself.}
An optimization cached attention execution information, saving preparation work but sometimes reusing information invalidated by a new request.
Repeated timing on a fixed prompt passed, while a sequence of different requests could hang.
We verified a repair that refreshed this information after each new prefill and added varied-request checks to the speed test available to agents.
In a later run, an agent encountered the hang through this check and implemented the repair.
Here, improving the evaluation made a previously hidden failure actionable for the agent.

\textbf{(4) Task scores alone cannot certify an engine; the code must also be inspected.}
Even with generation-based evaluation and termination checks, some final engines of the main runs computed only part of the output vocabulary.
GPT Chain A omitted about 39\% of vocabulary rows, while its GSM8K accuracy differed from vanilla by only $-1.36$ percentage points.
The reported tasks use English prompts and numeric answers, so they exercise limited vocabulary coverage; this candidate also retained EOS and could terminate normally.
Code inspection exposed this reward hacking against the intended full-vocabulary task~\citep{rewardhacking}.
The audit in Appendix~\ref{implementation-audit} records the restrictions; the task scores alone had not detected them.

The revisions and the audit above were made by us, not by the agents. A natural next step is to let the agents revise the evaluation as well, so that the objective evolves together with the engine.

\hypertarget{limitations-and-conclusion}{%
\section{Limitations and Conclusion}\label{limitations-and-conclusion}}

The study covers one engine, a single-request throughput objective, and two target models; the full organization and context comparisons are on Qwen. Each condition has one group run, and the Qwen speculative-decoding history contains two locally developed implementations that accelerate decoding and two adoptions. The agents use different vendor CLIs, and CLI revisions changed during the campaign, so observed behavior includes tooling effects. Development workloads fix generation length, while natural-length timing covers a small task workload and no concurrent serving load. Reported post-validation covers two task benchmarks and selected implementation properties; it does not establish unrestricted functional equivalence. Final task accuracy uses data excluded from development feedback, but the two numeric-answer tasks provide limited capability coverage.

SEIS shows agents coordinating changes across a complete inference engine over sustained autonomous work. Its best engine reaches 3.27\ensuremath{\times} vanilla throughput on natural requests, with 3.08\ensuremath{\times} retained when speculation is disabled. The histories connect this result to accumulated kernel and runtime improvements and to selective integration of a peer's speculative-decoding policy. Other final engines expose both gaps in development checks and accuracy warnings that do not prevent inheritance. The systems and their records therefore show how useful engineering accumulates, how implementations combine, and how development scores can diverge from delivered capabilities.

\clearpage
\subsection*{AI use statement}
Generative AI agents are experimental subjects in this study and are evaluated as part of the research methodology. During the experiments, the agents are tasked with optimizing inference engine code. Separately, generative AI tools were used to assist with English-language editing for readability, development and refinement of analysis code, and consistency checks.

\subsection*{Ethics statement}
The experiments use public model weights and benchmark data. No human-subject data were collected for this study. 

\subsection*{Reproducibility statement}
Appendices specify the workloads, development gates, final-engine selection, accuracy grading, and performance aggregation. Tables and data figures derive from consolidated session records. 

\bibliography{references}
\bibliographystyle{iclr2027_conference}

\clearpage
\appendix

\hypertarget{full-qwen-results-and-independent-starts}{%
\section{Full Qwen Results and the Branching Reference}\label{full-qwen-results-and-independent-starts}}

\begin{table}[htbp]
\caption{All Qwen endpoints: the last development-gate-passing engine of each chain. Speedups are factors relative to vanilla. Accuracy differences are percentage points on the 660-item GSM8K test split and retrieval; \textbf{!} marks an observed difference outside \ensuremath{\pm}3 points.}\label{tab:A1}
\centering
\footnotesize
\renewcommand{\arraystretch}{1.15}
\begin{tabular}{l l r r r r}
\toprule
condition & chain & development & held-out & GSM8K \ensuremath{\Delta} & retrieval \ensuremath{\Delta} \\
\midrule
\multicolumn{6}{l}{\textit{Opus}} \\
Chain & A & 4.94 & 6.11 & \ensuremath{-}0.15 & \ensuremath{-}1.48 \\
Chain & B & 2.88 & 2.71 & +0.91 & \ensuremath{-}28.89 \textbf{!} \\
Chain & C & 3.54 & 3.31 & +0.91 & +0.37 \\
Sync-3 & A & 3.38 & 3.12 & +1.52 & \ensuremath{-}0.37 \\
Sync-3 & B & 3.39 & 3.15 & +1.06 & 0.00 \\
Sync-3 & C & 3.26 & 3.11 & +0.30 & \ensuremath{-}1.48 \\
Sync-9 & A & 6.28 & 8.54 & +1.06 & \ensuremath{-}0.37 \\
Sync-9 & B & 5.50 & 7.69 & +0.91 & \ensuremath{-}1.11 \\
Sync-9 & C & 5.83 & 8.54 & \ensuremath{-}1.52 & \ensuremath{-}1.48 \\
Chain-128k & A & 2.93 & 2.73 & +1.06 & \ensuremath{-}0.37 \\
Chain-128k & B & 2.93 & 2.75 & +0.91 & \ensuremath{-}0.74 \\
Chain-128k & C & 3.01 & 2.79 & +1.21 & \ensuremath{-}1.11 \\
\addlinespace[3pt]\multicolumn{6}{l}{\textit{GPT}} \\
Chain & A & 2.43 & 2.37 & \ensuremath{-}1.36 & \ensuremath{-}0.74 \\
Chain & B & 1.84 & 1.82 & \ensuremath{-}1.21 & +1.11 \\
Chain & C & 2.13 & 2.07 & \ensuremath{-}1.36 & \ensuremath{-}2.22 \\
Sync-3 & A & 2.96 & 2.76 & +0.30 & 0.00 \\
Sync-3 & B & 2.95 & 2.84 & \ensuremath{-}0.15 & +0.74 \\
Sync-3 & C & 2.89 & 2.71 & \ensuremath{-}1.21 & \ensuremath{-}0.74 \\
Sync-9 & A & 2.22 & 2.16 & \ensuremath{-}2.58 & 0.00 \\
Sync-9 & B & 2.14 & 2.17 & \ensuremath{-}2.42 & \ensuremath{-}29.63 \textbf{!} \\
Sync-9 & C & 2.21 & 2.16 & \ensuremath{-}2.42 & 0.00 \\
Chain-128k & A & 2.23 & 2.09 & \ensuremath{-}4.39 \textbf{!} & +2.59 \\
Chain-128k & B & 2.39 & 0.77 & \ensuremath{-}4.55 \textbf{!} & \ensuremath{-}2.59 \\
Chain-128k & C & 1.94 & 1.88 & \ensuremath{-}4.39 \textbf{!} & +2.96 \\
\bottomrule
\end{tabular}
\end{table}

\begin{figure}[!t]
\hypertarget{fig:A1}{%
\centering
\includegraphics{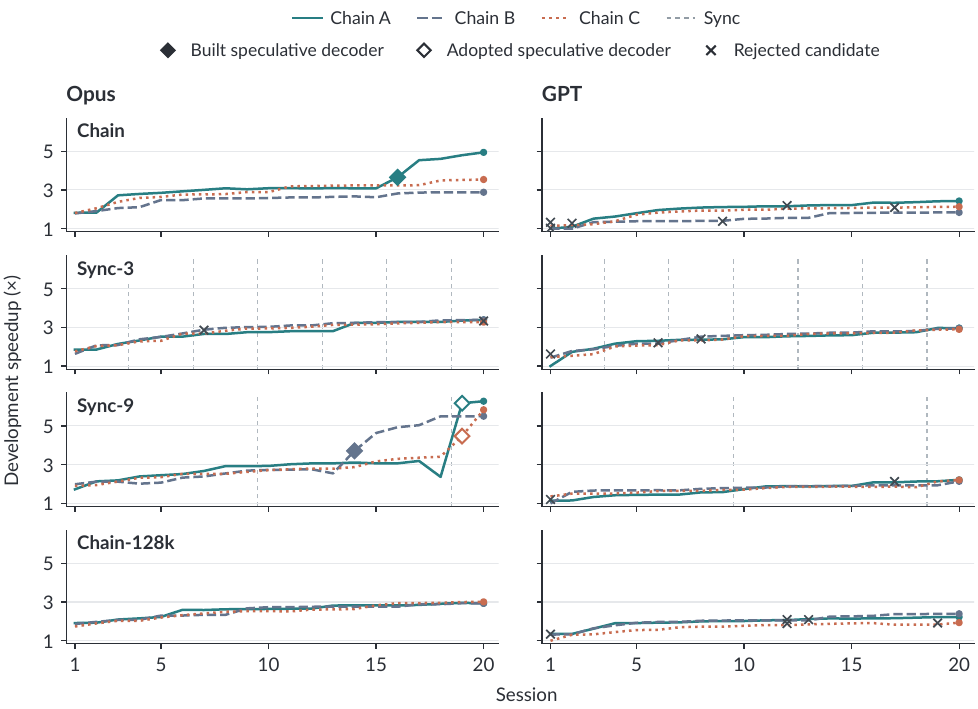}
\caption{Retained development speedup along every Qwen chain, with one common vertical scale. Chain colors and diamond markers match Figure~\ref{fig:6}: filled diamonds indicate speculative-decoder builds and open diamonds adoptions. Crosses show rejected candidates at their measured candidate scores; rejected or unscored sessions leave the retained state unchanged. Vertical dashed lines mark syncs.}\label{fig:A1}
}
\end{figure}

\textbf{Branching reference.} The reference uses first sessions from the large-window Chain, Sync-3, and Sync-9 conditions, separately for each agent: nine observed starts per agent. All nine Opus candidates pass the development gates; five of the nine GPT candidates pass. A rejected candidate contributes the vanilla engine (1.00\ensuremath{\times}), reflecting the engine retained after spending that session's budget. For each k from one to nine, the empirical best-of-k value is the average maximum over all k-element subsets of these outcomes. It is computed without replacement and is not a prediction about unobserved independent runs. The three Chain-128k first sessions per agent are kept in the context comparison and are not pooled into this reference.

\textbf{Reporting scope.} One original chain in the Opus Chain-128k condition could not complete post-validation and was replaced under a rule recorded before its replacement began. Its endpoint reached 2.47\ensuremath{\times} and built no working speculative decoder; four chains were attempted to obtain the three reported in that condition. The original plan emphasized the fastest quality-validated engine per condition. We report retained endpoints to expose final delivery; these are not an exhaustive search over earlier quality-validated checkpoints.

\hypertarget{speculative-decoding-implementation-and-engineering-history}{%
\section{Speculative Decoding: Implementation and Engineering History}\label{speculative-decoding-implementation-and-engineering-history}}

\textbf{Algorithm and integration.} The evolved speculative decoder uses n-gram lookup over the prompt and generated history, followed by verification against the candidate engine's greedy predictions. The policy ranks candidate continuations by backward context match, estimates token yield by context bucket, and selects a verification width using measured costs. The execution layer reserves KV slots, commits the accepted prefix, and pipelines verification with emission of the previous accepted tail. Multi-row INT8 kernels and shared-KV attention support verification widths of eight and sixteen in the best engine. These mechanisms implement established speculative-decoding and prompt-lookup ideas \citep{specdecode,promptlookup}; no new decoding-algorithm guarantee is claimed.

\textbf{Token yield.} The offline replay statistic is \emph{output tokens per verification step}, including the ordinary next token produced by a step. It is not the probability of accepting a draft token. With an eight-token draft and the replay policy specified in the analysis code, the geometric mean across the nine development shapes is 2.30 tokens/step. Two shapes have baseline distinct-4 below 0.3; their geometric mean is 4.21, versus 1.94 for the other seven. The corresponding held-out geometric mean is 3.46. These are replay estimates on vanilla outputs, not online measurements of the final adaptive policy.

The erroneous scripts compare a draft token with the output position one token ahead of the correct verification position. The resulting estimates near one token per step are recorded as reasons to abandon the technique. Correct replay changes that assessment. The availability of a profitable implementation also depends on the cost of multi-row verification, which changes with the engine's kernels. This history supports the process account in Section~\ref{accumulation-across-sessions} without assigning a separate causal contribution to every preceding edit.

\textbf{Builds and adoptions.} Opus Chain A completes a speculative implementation that accelerates decoding in session 16 and extends it in session 17. Opus Sync-9 B prepares the path in session 13 and rebuilds the slower prototype into an accelerating implementation in session 14. A and C adopt B's design after the sync following session 18. Source comparison finds 181 matching lines among A's 637 nonblank code lines, excluding comment-only lines and standalone string expressions such as docstrings. Matching preserves indentation and tests membership in B's corresponding code lines. Shared coding idioms can contribute to this count; the reports establish the adoption history, while the overlap corroborates it.

\textbf{Policy details and inherited judgments.} The policy maintains token-yield estimates by backward-match-length bucket and compares them with measured eight- and sixteen-row verification costs. It also scores candidate continuations against tokens from ordinary decoding, allowing updates without always taking a speculative step. Reports can lose useful counterevidence: Sync-3 B measures a yield of 2.07 tokens per step in session 13, but omits it from the session report; subsequent reports return to the belief that the generation interface blocks the technique. The build counts above identify implementations that deliver acceleration; earlier slower prototypes and A's session-18 implementation that emits only one token per verification step are treated as unsuccessful attempts.

\textbf{Verification coverage.} Development teacher-forced checks exercise the ordinary single-token path. We additionally capture 464 logit rows from actual eight- and sixteen-row verification steps, then replay the same prefixes with vanilla and with the optimized ordinary path. Contexts range from 64 to 2,058 tokens. Widths are selected for coverage, and two requests inject incorrect drafts. Table~\ref{tab:B1} pools the comparisons by width; the PPL ratio uses each reference row's argmax as its target. The optimized single-token path agrees on every top-1 prediction in this sample. Six controlled state cases check first and middle rejection, full acceptance, EOS within an accepted prefix, output-budget limits, and ignoring EOS. Runtime checks cover 421 verification steps, and an interleaved-request repeat confirms identical output after reset. In eight ordinary runs, all 280 observed speculative input prefixes agree with the returned sequence up to EOS or the request budget. These finite checks validate exercised paths without proving unrestricted functional equivalence.

\begin{table}[htbp]
\caption{Matched-prefix checks of the best engine's actual verification paths. Metrics are pooled over the indicated positions. Reference paths decode one token at a time.}\label{tab:B1}
\centering
\footnotesize
\renewcommand{\arraystretch}{1.15}
\begin{tabularx}{\linewidth}{Y r r r r r}
\toprule
reference & verify rows & positions & mean KL & top-1 (\%) & PPL ratio \\
\midrule
vanilla & 8 & 112 & 0.001605 & 98.21 & 1.0048 \\
vanilla & 16 & 352 & 0.001172 & 98.86 & 1.0001 \\
optimized ordinary & 8 & 112 & 0.000354 & 100.00 & 1.0003 \\
optimized ordinary & 16 & 352 & 0.000280 & 100.00 & 1.0007 \\
\bottomrule
\end{tabularx}
\end{table}

\hypertarget{session-records-and-evolution-patterns}{%
\section{Session Records and Evolution Patterns}\label{session-records-and-evolution-patterns}}

\begin{table}[htbp]
\caption{Main Qwen session records. Each condition has sixty indexed sessions. Passing and rejected counts refer to development gates; an unscored session contributes no candidate verdict. Durations describe recorded session work, not a total monetary-cost comparison.}\label{tab:C1}
\centering
\footnotesize
\renewcommand{\arraystretch}{1.15}
\begin{tabular}{l r r r r}
\toprule
condition & passing & rejected & unscored & median minutes \\
\midrule
\multicolumn{5}{l}{\textit{Opus}} \\
Chain & 60 & 0 & 0 & 109.2 \\
Sync-3 & 57 & 2 & 1 & 83.2 \\
Sync-9 & 58 & 0 & 2 & 115.0 \\
Chain-128k & 58 & 0 & 2 & 120.0 \\
\addlinespace[3pt]\multicolumn{5}{l}{\textit{GPT}} \\
Chain & 54 & 6 & 0 & 62.0 \\
Sync-3 & 57 & 3 & 0 & 49.6 \\
Sync-9 & 58 & 2 & 0 & 52.8 \\
Chain-128k & 54 & 5 & 1 & 100.8 \\
\bottomrule
\end{tabular}
\end{table}

Across the main matrix, Opus's median recorded session lasts 112.5 minutes and GPT's 62.7 minutes. The two-hour cap is reached in 103/240 and 35/240 sessions, respectively. Command histories also differ: the median numbers of commit commands are four and thirteen. These measurements describe the observed runs and do not define fixed agent traits.

Within GPT Sync-3, the median increase in development speedup is 0.112 immediately after a sync and 0.025 in other eligible sessions (18 and 36 observations). The corresponding Opus Sync-3 medians are 0.050 and 0.037. Sessions in a chain are dependent, and Sync-3 chains exchange information; these counts are descriptive and are not independent replications of a synchronization policy.

\begin{figure}
\hypertarget{fig:C1}{%
\centering
\includegraphics{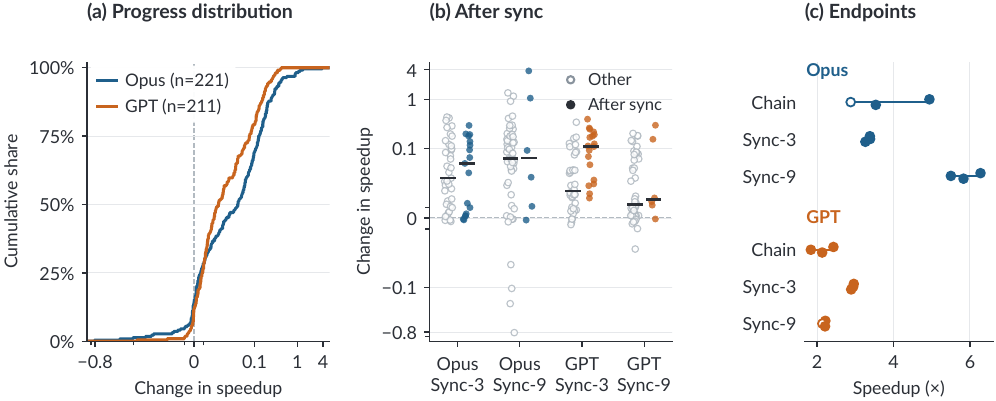}
\caption{Descriptive evolution statistics. (a) Cumulative distributions of all changes between consecutive passing engines within each chain. (b) Changes after sync and in other eligible sessions; horizontal marks are medians. Change axes are symmetric-log, linear within \ensuremath{\pm}0.05. (c) Three endpoints per large-window condition with min--max spans; open markers indicate observed task-accuracy differences outside \ensuremath{\pm}3 points. Sessions and synchronized chains are dependent observations.}\label{fig:C1}
}
\end{figure}

\textbf{Compaction.} Opus's retained Chain-128k condition records 242 compactions across sixty sessions. Summaries replace contexts of roughly 130--139k tokens with substantially shorter reports, and 43 sessions reach the two-hour cap. GPT's rollout logs also record compaction in all sixty Chain-128k sessions: 287 events in total, including seven in the unscored session. No claim is made that a compaction necessarily causes the next measured regression.

\textbf{Context comparison.} Opus reaches the two-hour cap in 43 of its sixty Chain-128k sessions, compared with 23 in the large-window Chain condition. Its implementation attempt in Chain-128k starts only in session 20 and stops at the cap. GPT's development-speedup ranges overlap across budgets: 1.94--2.39\ensuremath{\times} in Chain-128k and 1.84--2.43\ensuremath{\times} in the large-window Chain condition; neither produces a working speculative decoder. Compaction changes both the available history and the remaining work time, which these runs do not separately control.

\textbf{Measurement variation.} Thirty-seven consecutive passing-session pairs in the recorded analysis leave the source tree unchanged. Their development-score differences range from \ensuremath{-}0.010 to +0.027 in speedup units, with median absolute difference 0.004. This supplies an empirical scale for interpreting small changes in those runs, not a universal noise bound. Appendix E separately reports repeat measurements of the best engine and its ablation.

\hypertarget{evaluation-protocol}{%
\section{Evaluation Protocol}\label{evaluation-protocol}}

\textbf{Development workloads.} The nine development shapes cross three content types (natural language, mathematics, and code) with input/output lengths 256/256, 512/128, and 1,024/256. The eighteen held-out shapes use input lengths 128 and 2,048 and output lengths 128, 512, and 1,024 for each content type. The same target-specific vanilla implementation provides the speed and numerical references. Timing reports the median of three runs per shape; aggregate speedup is the geometric mean of per-shape throughput ratios.

\textbf{Development gates.} Numerical checks use full-vocabulary logits conditioned on the vanilla token sequence. KL is computed in the vanilla-to-candidate direction and averaged over checked token positions. Perplexity ratio must not exceed 1.05, mean KL must not exceed 0.05, and top-1 agreement must be at least 0.88. Separate free-generation runs check agreement between timing and instrumentation modes. Four EOS-enabled termination prompts must finish before their caps. The GSM8K feedback bit is recorded and shown to the next session, but it is not a gate.

\textbf{Task accuracy.} We partition GSM8K's released evaluation items using a fixed seed into 659 development-feedback items and 660 final-test items. All reported final GSM8K task accuracies use the latter, with greedy generation and the saved prompt format. The synthetic retrieval test has thirty items in each combination of filler length (512, 1,024, or 2,048 tokens) and fact position (10 \%, 50 \%, or 90 \%), giving 270 items. Four named access-code facts are inserted into WikiText filler \citep{wikitext}, and the model must return the requested four-digit number. The Qwen prompts including instructions contain 592--2,133 tokens and allow up to 32 output tokens. This tests longer-input retrieval, distinct from the agent's context-window experiment.

Both tasks use the recorded numeric-answer grader. For a generation that reaches its cap, extraction ends at the first answer-closing marker on GSM8K, when present, or the first numeric response on retrieval; other outputs are unchanged. This correction changes no item-level grade for the current Qwen endpoints or vanilla. Paired comparisons use the same items for both engines. The \ensuremath{\pm}3-point margin describes observed differences; statistical equivalence additionally requires both one-sided tests at level 0.05.

\textbf{Protocol scope.} Earlier pilot failures informed the main harness, including its termination probe and the handling of stalled accuracy checks. Those pilot cases are identified as such in Section~\ref{reward-hacking-and-validation-failures} and do not enter the 480-session main matrix. All experiments reported as main conditions use the same development thresholds and inheritance rule. Protected benchmark and interface files are restored before scoring, and scoring runs outside the agent's process.

\hypertarget{timing-generalization-and-accuracy-details}{%
\section{Timing, Generalization, and Accuracy Details}\label{timing-generalization-and-accuracy-details}}

\textbf{Timing conditions.} Measurements retain warm-up and time synchronous local engine calls, including prefill and generation, with one request at a time. Development timing uses mini-sglang's default prefix cache; natural-length and production comparisons disable cross-request prefix reuse. Configurations with a shared draft cache clear it between repetitions and use disjoint warm-up requests. Model loading, compilation, and profiling are outside throughput measurements. Latency statistics use the repetition with median aggregate throughput.

\textbf{Matched ablation.} The matched measurement freezes the vanilla and best-engine source trees and changes only the best engine's speculation switch between its two variants. Each configuration measures the same 100 prompts with EOS enabled and again with 256 forced output tokens, for three repetitions. Warm-up precedes timing, cross-request prefix caching is disabled, and the GPU is synchronized before and after each request. Throughput for each repetition is total emitted tokens divided by total request time. The fixed-256 workload in this ablation is not the nine-shape development grid.

\begin{table}[htbp]
\caption{Matched ablation on 100 GSM8K prompts. Throughput is the median of three repetitions; brackets show the minimum and maximum, not a confidence interval. Each speedup divides the median throughput by the vanilla median under the same length rule.}\label{tab:E1}
\centering
\footnotesize
\renewcommand{\arraystretch}{1.15}
\begin{tabularx}{\linewidth}{Y >{\raggedright\arraybackslash}p{2.4cm} r >{\raggedright\arraybackslash}p{2.4cm} r}
\toprule
engine & natural tokens/s {[}range{]} & speedup & fixed-256 tokens/s {[}range{]} & speedup \\
\midrule
vanilla & 594 {[}583--594{]} & 1.00\ensuremath{\times} & 599 {[}599--599{]} & 1.00\ensuremath{\times} \\
evolved, speculation off & 1,831 {[}1,725--1,837{]} & 3.08\ensuremath{\times} & 1,866 {[}1,846--1,876{]} & 3.12\ensuremath{\times} \\
evolved, speculation on & 1,942 {[}1,833--1,943{]} & 3.27\ensuremath{\times} & 2,317 {[}2,317--2,318{]} & 3.87\ensuremath{\times} \\
\bottomrule
\end{tabularx}
\end{table}

Natural-length outputs are identical across the three repetitions within each configuration. Final task accuracy is evaluated separately on the 660-item test split.

The matched measurement uses PyTorch 2.9.1+cu128 and CUDA 12.8, with one PyTorch CPU thread, on the same allocated H100 for all three variants.

\begin{table}[htbp]
\caption{Natural-request output lengths and latency on the matched 100-request workload. Outputs are identical across repetitions within each configuration. Latency statistics use the repetition with median aggregate throughput; configurations generate their own answers.}\label{tab:E2}
\centering
\footnotesize
\renewcommand{\arraystretch}{1.15}
\begin{tabularx}{\linewidth}{Y r r r r}
\toprule
engine & output tokens & mean length & mean latency (ms) & median latency (ms) \\
\midrule
vanilla & 17,541 & 175.41 & 295.47 & 257.55 \\
evolved, speculation off & 17,614 & 176.14 & 96.19 & 90.54 \\
evolved, speculation on & 17,319 & 173.19 & 89.20 & 86.75 \\
\bottomrule
\end{tabularx}
\end{table}

\textbf{Profiling.} A separate CUDA trace profiles the first of those prompts with 128 forced output tokens. Figure~\ref{fig:3} groups GPU kernels by function and reports their summed duration and invocation count per emitted token, including prefill, for vanilla and the full evolved engine. With speculation off, the evolved engine records 775 \ensuremath{\mu}s of summed kernel time and 145 kernel invocations per output token. Fused operations are assigned to their encompassing kernel group. Device work may overlap, and profiling adds overhead, so these totals are neither an additive wall-clock decomposition nor the source of the throughput numbers.

\begin{figure}
\hypertarget{fig:E1}{%
\centering
\includegraphics{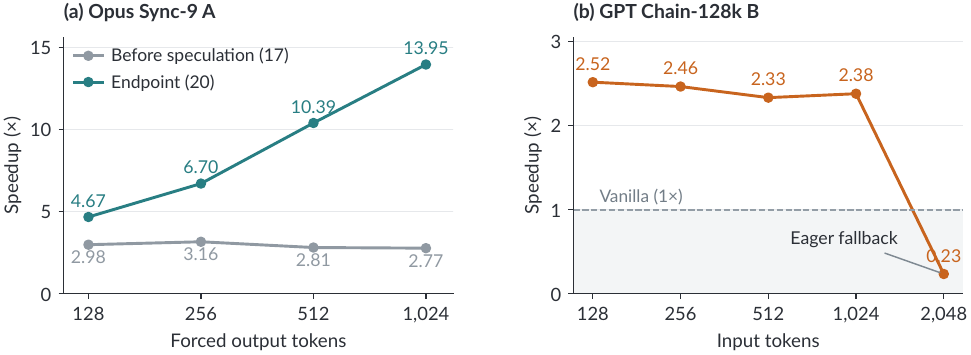}
\caption{Workload dependence. Points report geometric-mean speedup over recorded shapes with the indicated length. (a) Opus Sync-9 A before speculation and at its endpoint; the checkpoints differ in several implementation changes. (b) GPT Chain-128k B loses performance when longer inputs exceed its CUDA-graph coverage and trigger eager execution. The dashed line is vanilla throughput; the shaded region indicates a slowdown. Each length group contains its observed combinations of prompt types and other sequence lengths.}\label{fig:E1}
}
\end{figure}

\textbf{Production references.} We use vLLM 0.19.1, SGLang 0.5.10, and TensorRT-LLM 1.0.0 in isolated environments on the same H100 allocation as the matched SEIS measurements. Their PyTorch versions are 2.10.0, 2.9.1, and 2.7.1, respectively, with CUDA 12.8 runtimes. Public settings restrict execution to one active request and a 4,096-token context limit. The search considers BF16, supported FP8 and weight-only INT8 modes, n-gram settings spanning five to sixteen draft tokens, and available execution or attention backends. We time 32 development requests three times; potentially competitive configurations must have observed accuracy within \ensuremath{\pm}3 points of vanilla on the 659 development items. A configuration slower than an already qualified choice needs no further quality measurement. Final configuration selection uses only development data; fixed configurations are evaluated on 100 timed requests, 660 GSM8K test items, and 270 retrieval items.

\begin{table}[htbp]
\caption{Selected production configurations and final accuracy differences in percentage points relative to vanilla. Decode CUDA graphs are enabled. SGLang uses eager prefill; TensorRT-LLM n-gram decoding uses a private per-request pool and disables overlap scheduling.}\label{tab:E3}
\centering
\footnotesize
\renewcommand{\arraystretch}{1.15}
\begin{tabularx}{\linewidth}{l Y r r}
\toprule
framework & selected settings & GSM8K \ensuremath{\Delta} & retrieval \ensuremath{\Delta} \\
\midrule
vLLM & BF16; no speculation & +0.61 & \ensuremath{-}0.74 \\
SGLang & BF16; FlashInfer; no speculation & +0.30 & \ensuremath{-}0.37 \\
TensorRT-LLM & BF16; PyTorch backend; n-gram 5 & +0.76 & \ensuremath{-}1.11 \\
\bottomrule
\end{tabularx}
\end{table}

\textbf{Best-engine accuracy.} On the 660-item GSM8K test split, vanilla answers 379 items correctly and the best engine 386: 57.42 \% versus 58.48 \%, or +1.06 points. The paired normal-approximation 90 \% confidence interval is {[}\ensuremath{-}0.92, +3.04{]}. Two-sided equivalence within \ensuremath{\pm}3 points is not established; the interval extends just beyond the positive boundary. On retrieval, the counts are 238/270 and 237/270: 88.15 \% versus 87.78 \%, or \ensuremath{-}0.37 points, and paired equivalence is established. These statistical readings are distinct from the observed differences reported in the main tables.

The test split was fixed before optimization and never supplied feedback. All Qwen and Llama accuracy summaries use these same test item IDs, paired with each target model's vanilla outputs.

\FloatBarrier

\hypertarget{implementation-audit}{%
\section{Implementation Audit}\label{implementation-audit}}

The audit compares endpoint code with vanilla at the level of numerical representation, attention, model execution, scheduling, and output generation. Common useful changes include INT8 or lower-bit linear kernels, split-KV attention, fused normalization and activation, device-side argmax, and CUDA-graph restructuring. These technique families are established in inference systems; this paper studies their autonomous implementation and integration rather than claiming each as a new technique.

\begin{figure}[!h]
\hypertarget{fig:F1}{%
\centering
\includegraphics{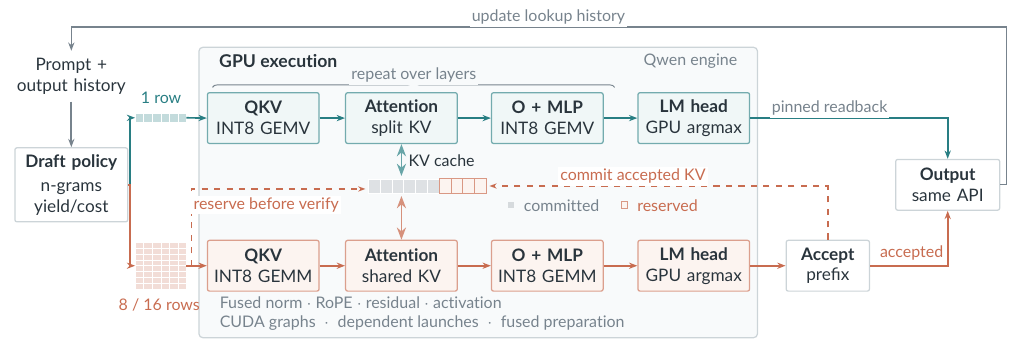}
\caption{Execution paths of the best engine. Teal denotes ordinary single-token decoding; coral denotes multi-token verification. Rows represent token positions within one request. Both paths use the optimized model computation and shared KV storage. Dashed arrows denote KV reservation and commitment; accepted tokens return through the existing generation interface. The schematic groups repeated Transformer operations.}\label{fig:F1}
}
\end{figure}

\begin{table}[htbp]
\caption{Changes in GPT endpoints that restrict computation or depend on evaluation mode. Development speedups describe the complete endpoints and are not speedups attributable to the individual listed change.}\label{tab:F1}
\centering
\footnotesize
\renewcommand{\arraystretch}{1.15}
\begin{tabularx}{\linewidth}{l l r >{\raggedright\arraybackslash}p{2.3cm} >{\raggedright\arraybackslash}p{2.3cm} Y}
\toprule
condition & chain & development & omitted output rows & EOS-dependent execution & other observed changes \\
\midrule
Chain & A & 2.43\ensuremath{\times} & 59,872 & --- & FP8 KV on four layers \\
Chain & B & 1.84\ensuremath{\times} & 51,910 & deferred multi-step execution & --- \\
Chain & C & 2.13\ensuremath{\times} & --- & --- & skips layer 23's attention at selected context lengths \\
Sync-3 & A & 2.96\ensuremath{\times} & 19,840 & --- & fitted RMS denominator \\
Sync-3 & B & 2.95\ensuremath{\times} & 53,072 & --- & --- \\
Sync-3 & C & 2.89\ensuremath{\times} & 53,072 & --- & --- \\
Sync-9 & A & 2.22\ensuremath{\times} & 49,152 & multi-step graph replay & --- \\
Sync-9 & B & 2.14\ensuremath{\times} & at least 49,152 & multi-step graph replay & --- \\
Sync-9 & C & 2.21\ensuremath{\times} & 49,152 & multi-step graph replay & --- \\
Chain-128k & A & 2.23\ensuremath{\times} & --- & --- & INT6 rows and INT4 remainder, full vocabulary \\
Chain-128k & B & 2.39\ensuremath{\times} & --- & --- & eager fallback beyond graph coverage \\
Chain-128k & C & 1.94\ensuremath{\times} & dynamic selection above row 52,429 & grouped replay & probabilistic row selection \\
\bottomrule
\end{tabularx}
\end{table}

The best engine retains the full output head and attention computation. In the restricted-head GPT example, uncomputed logits receive fixed fill values; retaining the forced token alone does not guarantee agreement over the full vocabulary. Passing verdicts on the development inputs demonstrate the limited coverage of those checks.

\textbf{Rejected design choices.} In the best engine's chain, group-wise INT4 weights fail the top-1 agreement gate, FP8 decode weights provide no gain, and a 32-row verification graph is slower than the retained widths. These tests document local selection among alternative numerical representations and verification widths. Multi-row kernels use tensor cores, attention shares KV reads across rows, and producer kernels emit partial sums used by the next RMS normalization; the detailed paths appear in Figure~\ref{fig:F1}.

\textbf{Audit and accuracy counts.} The nine restricted-head endpoints and the attention-skipping endpoint in Table~\ref{tab:F1}, together with the five endpoints outside the final-task \ensuremath{\pm}3-point margin, identify thirteen distinct endpoints. GPT Sync-9 B and GPT Chain-128k C appear in both the audit and accuracy groups. Five of the thirteen have a recorded development GSM8K FAIL: GPT Chain A/B/C and GPT Chain-128k B/C. The bit is informational and does not invalidate the numerical and termination verdict. These counts use the retained endpoints and the two reported final tasks; no historical retention decision is changed by the audit.

\FloatBarrier

\hypertarget{sequential-optimization-on-llama-3.2-1b-instruct}{%
\section{Branching Chain Optimization on Llama-3.2-1B-Instruct}\label{sequential-optimization-on-llama-3.2-1b-instruct}}

Llama-3.2-1B-Instruct \citep{llama32} uses the same engine, H100 platform, session cap, large context settings, development shapes, and inheritance rule. Target-specific tokenization, architecture information, and vanilla reference outputs are regenerated. Opus and GPT each run the Chain condition with three chains of twenty sessions. This appendix reports those two completed conditions; the organization and context comparisons in the main paper remain Qwen experiments.

\begin{table}[htbp]
\caption{Llama Chain endpoints, all at session 20. Speedups are factors relative to the Llama vanilla engine; accuracy differences are percentage points on the 660-item GSM8K test split and retrieval. Bold marks a difference outside \ensuremath{\pm}3 points; \(\dagger\) marks speculation.}\label{tab:G1}
\centering
\small
\renewcommand{\arraystretch}{1.15}
\begin{tabular}{c r r r r @{\hspace{12pt}} r r r r}
\toprule
 & \multicolumn{4}{c}{Opus} & \multicolumn{4}{c}{GPT} \\
\cmidrule(lr){2-5}\cmidrule(lr){6-9}
chain & dev. & held-out & GSM8K \ensuremath{\Delta} & retrieval \ensuremath{\Delta} & dev. & held-out & GSM8K \ensuremath{\Delta} & retrieval \ensuremath{\Delta} \\
\midrule
A & 5.66\rlap{\(^\dagger\)} & 7.02 & \ensuremath{-}1.36 & 0.00 & 2.80 & 2.64 & \textbf{\ensuremath{-}3.48} & \textbf{\ensuremath{-}11.11} \\
B & 5.82\rlap{\(^\dagger\)} & 6.53 & \ensuremath{-}1.67 & 0.00 & 2.53 & 2.47 & \ensuremath{-}2.88 & \ensuremath{-}0.74 \\
C & 2.63 & 2.59 & \ensuremath{-}1.06 & \ensuremath{-}0.74 & 2.50 & 2.21 & \ensuremath{-}0.91 & \textbf{\ensuremath{-}11.11} \\
\bottomrule
\end{tabular}
\end{table}

Vanilla accuracies are 48.79 \% on the GSM8K test split and 99.63 \% on retrieval. All three Opus endpoints and GPT B have observed differences within \ensuremath{\pm}3 points on both sets; paired GSM8K equivalence is not established for these endpoints. GPT A and C each lose 11.11 retrieval points and err on the same 31 items. An earlier GPT A checkpoint at session 6 has a GSM8K test difference of \ensuremath{-}0.30 points and is timed separately below.

Opus A and B implement prompt-lookup speculative decoding at sessions 14 and 13, respectively. Their development speedups change from approximately 2.61\ensuremath{\times} to 4.25\ensuremath{\times} and 2.43\ensuremath{\times} to 3.51\ensuremath{\times} in those sessions. Opus C improves kernels and execution without building a speculative decoder, ending at 2.63\ensuremath{\times}. None of the GPT chains implements speculation. These records add two examples of structural changes appearing after preceding optimization, without establishing a required discovery time.

\begin{table}[htbp]
\caption{Archived natural-length Llama throughput on 100 GSM8K requests, with EOS enabled and prefill included. Factors use unrounded throughput. The GPT entry is an earlier checkpoint; its three endpoints were not timed under this protocol.}\label{tab:G2}
\centering
\footnotesize
\renewcommand{\arraystretch}{1.15}
\begin{tabularx}{\linewidth}{Y c r r}
\toprule
engine & session & tokens/s & speedup over vanilla \\
\midrule
vanilla mini-sglang & --- & 629 & 1.00\ensuremath{\times} \\
Opus A & 20 & 1,820 & 2.90\ensuremath{\times} \\
Opus B & 20 & 1,966 & 3.13\ensuremath{\times} \\
Opus C & 20 & 1,657 & 2.64\ensuremath{\times} \\
GPT A & 6 & 1,333 & 2.12\ensuremath{\times} \\
\bottomrule
\end{tabularx}
\end{table}

Opus B reaches 3.13\ensuremath{\times} natural-length speedup, compared with 5.82\ensuremath{\times} development speedup. Opus C reaches 2.64\ensuremath{\times} at natural length and 2.63\ensuremath{\times} on development shapes. Substantial kernel and runtime improvements therefore remain useful without speculation. The cross-chain comparison does not isolate speculation's contribution because the other implementations differ.

\hypertarget{reproduction-and-data-consistency}{%
\section{Reproduction and Data Consistency}\label{reproduction-and-data-consistency}}

The analysis uses saved working trees, session reports, transcripts, per-shape scores, development verdicts, and post-validation outputs. The collection records preserve missing scores explicitly. Endpoint selection uses the last passing working tree, and its speed and accuracy measurements are associated with that same version. The Qwen and Llama results are aggregated separately, and the Branching reference uses only the matching large-window Qwen conditions.

The manuscript data and figures are generated from the consolidated session records and the additional measurement outputs. The matched ablation records source hashes, workload identifiers, runtime versions, repetition-level throughput, per-request outputs, and CUDA profiles. Accuracy-subset statistics are recomputed from saved responses using the original item split and grader. The protocol, selection rules, benchmark construction, and numerical aggregation are specified in Appendices A--G; the recorded runs remain the basis for claims about these particular model and CLI versions.

\end{document}